\pdfoutput=1

\documentclass[11pt]{article}

\usepackage[]{EMNLP2023}

\usepackage{times}
\usepackage{latexsym}
\usepackage{enumitem}
\usepackage{graphicx}
\usepackage{multirow}
\usepackage{booktabs}
\usepackage{bbding}
\usepackage{diagbox}
\usepackage{utfsym}

\newcommand\snowman{\raisebox{-2pt}{\includegraphics[width=1.2em]{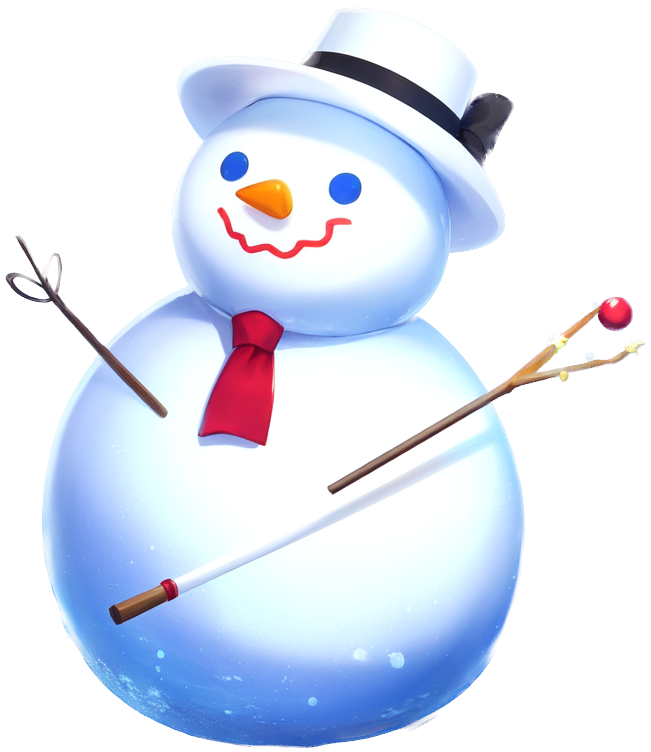}}}

\usepackage[T1]{fontenc}

\usepackage[utf8]{inputenc}

\usepackage{microtype}

\title{Snowman \snowman{}: A Million-scale Chinese Commonsense Knowledge Graph Distilled from Foundation Model}

\author{Jiaan Wang\textsuperscript{$\spadesuit$}, \ Jianfeng Qu\textsuperscript{$\spadesuit$}, \ Yunlong Liang\textsuperscript{$\diamondsuit$}, \ Zhixu Li\textsuperscript{$\clubsuit$}\\
\bf {An Liu\textsuperscript{$\spadesuit$}, \ Guanfeng Liu\textsuperscript{$\heartsuit$} and \ Xin Zheng\textsuperscript{$\Im$}} \\
\textsuperscript{$\spadesuit$}Soochow University \qquad  \textsuperscript{$\clubsuit$}Fudan University \qquad \textsuperscript{$\diamondsuit$}Beijing Jiaotong University	\\
\textsuperscript{$\heartsuit$}Macquarie University \qquad \textsuperscript{$\Im$}iFLYTEK Research \\
\texttt{jawang.nlp@gmail.com}, \texttt{jfqu@suda.edu.cn} \\
\texttt{yunlonliang@gmail.com}, \texttt{zhixuli@fudan.edu.cn}
}

\begin{document}
\maketitle
\begin{abstract}
Constructing commonsense knowledge graphs (CKGs) has attracted wide research attention due to its significant importance in cognitive intelligence. Nevertheless, existing CKGs are typically oriented to English, limiting the research in non-English languages. Meanwhile, the emergence of foundation models like ChatGPT and GPT-4 has shown promising intelligence with the help of reinforcement learning from human feedback.
Under the background, in this paper, we utilize foundation models to construct a Chinese CKG, named Snowman. Specifically, we distill different types of commonsense head items from ChatGPT, and continue to use it to collect tail items with respect to the head items and pre-defined relations. Based on the preliminary analysis, we find the negative commonsense knowledge distilled by ChatGPT achieves lower human acceptance compared to other knowledge. Therefore, we design a simple yet effective self-instruct filtering strategy to filter out invalid negative commonsense. Overall, the constructed Snowman covers more than ten million Chinese commonsense triples, making it the largest Chinese CKG. Moreover, human studies show the acceptance of Snowman achieves 90.6\%, indicating the high-quality triples distilled by the cutting-edge foundation model. We also conduct experiments on commonsense knowledge models to show the usability and effectiveness of our Snowman.\footnote{The codes and data will be released upon publication.}

\end{abstract}

\section{Introduction}

Collecting large-scale commonsense knowledge is a long-standing goal of cognitive intelligence~\cite{feigenbaum1984knowledge,lenat1995cyc}.
The commonsense knowledge graphs (CKGs), as a type of knowledge carrier, contain structured knowledge about everyday concepts and their properties~\cite{advanced}, facilitating various downstream tasks, such as question answering~\cite{tian-etal-2020-scene}, storytelling~\cite{Wang2022IncorporatingCK,Ammanabrolu_Cheung_Broniec_Riedl_2021}, dialogue generation~\cite{liang2020infusing,liang2022emotional} and summarization~\cite{liang-etal-2022-variational,liang2022summary,liang2023d,10.1145/3488560.3498405,Wang2022ASO,Wang2022ClidSumAB,Wang2023TowardsUM}.
Many efforts have been devoted to constructing large-scale CKGs through manual collecting~\cite{speer2017conceptnet,ATOMIC,mostafazadeh-etal-2020-glucose} or automatic extraction~\cite{10.1145/2556195.2556245,10.1145/3357384.3357955,ijcai2020p554,advanced,wang-etal-2022-cn}.

Nevertheless, current CKGs are typically limited to English, \emph{e.g.}, TransOMCS~\cite{ijcai2020p554}, ATOMIC~\cite{ATOMIC} and ATOMIC 2020~\cite{Hwang2020COMETATOMIC2O}. Though ConceptNet~\cite{speer2017conceptnet} is a multi-lingual CKG, most of its non-English parts are lexical knowledge and are limited in quantity as well as coverage~\cite{wang-etal-2022-cn}.
To go beyond English-centric CKGs, some recent studies construct Chinese CKGs in different manners. For example, C3KG~\cite{li-etal-2022-c3kg}, as a Chinese commonsense conversation knowledge graph, is constructed based on the translation of ATOMIC~\cite{ATOMIC,Hwang2020COMETATOMIC2O}.
As revealed by some work, directly translating resources suffers from the issues of cultural differences~\cite{wang-etal-2022-cn} and translationese~\cite{yu-etal-2022-beyond,Wang2022UnderstandingTI}, which may limit the precise and broad expression of native commonsense knowledge.

Recently, thanks to the development of pre-trained language models (PLMs) (\emph{e.g.}, T5~\cite{raffel2020exploring} and GPT-3~\cite{brown2020language}), which makes it possible to extract commonsense knowledge directly from PLMs.
\citet{west2021symbolic} show that the CKG built from PLMs can surpass the crowdsourced ones in both quantity and quality.
Under this background, CN-AutoMIC~\cite{wang-etal-2022-cn} is created through distilling Chinese commonsense knowledge from mT5-XXL (a popular multi-lingual PLM) with 13B parameters~\cite{xue-etal-2021-mt5}.
Though great success has been achieved, the unfiltered CN-AutoMIC only reaches 47.6\% human acceptance and limited coverage.
This is because mT5 is pre-trained with the general language modeling task (\emph{i.e.}, span corruption), and does not suffer from instruction tuning~\cite{wei2021finetuned} and reinforcement learning from human feedback (RLHF)~\cite{stiennon2020learning}.
The latest research including ChatGPT~\cite{ChatGPT} and GPT-4~\cite{OpenAI2023GPT4TR} foundation models indicates that instruction tuning and RLHF could help machines to better follow the human input and give the satisfied generation.
In detail, instruction tuning finetunes PLMs on a collection of datasets described via instructions, and can substantially improve the zero-shot ability of PLMs on unseen tasks~\cite{wei2021finetuned,Muennighoff2022CrosslingualGT}.
RLHF uses human preferences as rewards to fine-tune PLMs using reinforcement learning (\emph{e.g.}, PPO algorithm~\cite{Schulman2017ProximalPO}), and it could help PLMs to generate high-quality responses as humans expected, promoting the early appearance of artificial general intelligence (AGI)~\cite{bubeck2023sparks,Wang2023IsCA,Wang2023CrossLingualSV}.

In this paper, we leverage the success of foundation models to build a new Chinese CKG named \textbf{Snowman} (Chinese common\textbf{\underline{s}}ense k\textbf{\underline{now}}ledge graph distilled fro\textbf{\underline{m}} Ch\textbf{\underline{a}}tGPT fou\textbf{\underline{n}}dation model).
Specifically, we distill structured Chinese commonsense knowledge from ChatGPT~\cite{ChatGPT} (a representative cutting-edge foundation model that has been pre-trained with instruction tuning and RLHF) via prompting and filtering.
Following CN-AutoMIC~\cite{wang-etal-2022-cn}, we first use small-scale seeds to collect about 185.1K commonsense head items from ChatGPT, including voluntary items (\emph{e.g.}, \emph{PersonX learns to cook}), involuntary items (\emph{e.g.}, \emph{PersonX is attacked}) and state items (\emph{e.g.}, \emph{PersonX is excited}).
Then, for each head item, we obtain its corresponding tail items based on several mainstream pre-defined relations such as \emph{xWant} and \emph{xNeed}.
After that, we collect a large number (11.1 million) of commonsense knowledge triples, and we further conduct preliminary analysis on their quality.
We find that the collected negative knowledge that describes what is hindered by and what makes the thing cannot happen, such as ``\emph{Renting a luxury car is hindered by bad driving records}'' and ``\emph{Doing exercise every day is hindered by a critical illness}'', achieves lower human acceptance compared with others. This is because most of the world knowledge exists in a positive and affirmative form~\cite{molnar2000truthmakers,barker2012being}, resulting in a limited ability to generate negative commonsense knowledge from PLMs~\cite{hossain-etal-2022-analysis}.
To this end, we utilize a simple yet effective self-instruct filtering strategy to filter out low-quality negative commonsense knowledge in the preliminary collection.
Specifically, though ChatGPT struggles to generate negative knowledge, it can well determine the rationality of negative knowledge (making a judgment is easier than generation).
Therefore, we sample a number of vanilla negative knowledge triples, and utilize ChatGPT to provide labels of whether each negative triple is reasonable.
We use these boolean data to train a binary classifier to filter out low-quality negative knowledge triples.
Finally, our Snowman contains 185.1K unique head items, 5.4M unique tail items and 10.5M triples, making it the largest Chinese CKG.
Human studies show the acceptance of Snowman reaches 90.6\%, indicating its high quality and the effectiveness of our construction method.
The number of triples involved in Snowman is 9.7 times that of the previous largest Chinese CKG with the same level of quality (\emph{i.e.}, CN-AutoMIC$_{high}$~\cite{wang-etal-2022-cn} with 89.5\% human acceptance).
Moreover, we also conduct experiments on commonsense knowledge generation models, \emph{i.e.}, COMET~\cite{Bosselut2019COMETCT}, to demonstrate the usability and effectiveness of our Snowman.

We highlight our contributions as follows:
\begin{itemize}[leftmargin=*,topsep=0pt]
\item We construct Snowman, a new Chinese commonsense knowledge graph (CKG) with 10.5M high-quality triples, which is 9.2 times that of the previous largest Chinese CKG with the same level of quality. Thanks to the rapid development of foundation models, our Snowman satisfies people in quality and quantity.
\item To the best of our knowledge, we are the first to collect commonsense knowledge through distilling from a foundation model (ChatGPT) that is trained with instruction tuning and reinforcement learning from human feedback. Both training paradigms improve the data quality distilled by the foundation model.
\item To deal with the negative knowledge issue, we utilize a simple yet effective self-instruct strategy that does not need any human annotation and can effectively filter out low-quality negative knowledge triples.
\item We conduct human studies and experiments on commonsense knowledge models to give an in-depth understanding of the CKG distilled by the foundation model.
The results show the high quality and effectiveness of our Snowman.

\end{itemize}

\section{Data Construction}
In this section, we discuss the construction process of Snowman. We first distill different types of head items from ChatGPT based on a small number of seed items (\S~\ref{subsec:2.1}). Then, for each head item, we continue to distill ChatGPT to collect its corresponding tail items with respect to different pre-defined relations (\S~\ref{subsec:2.2}). Lastly, we find the negative knowledge triples generated by ChatGPT suffer from mixed quality, and we use a simple yet effective self-instruct filtering strategy to filter out low-quality negative knowledge triples in the preliminary distillation (\S~\ref{subsec:2.3}).

\begin{figure}[t]
\centerline{\includegraphics[width=0.45\textwidth]{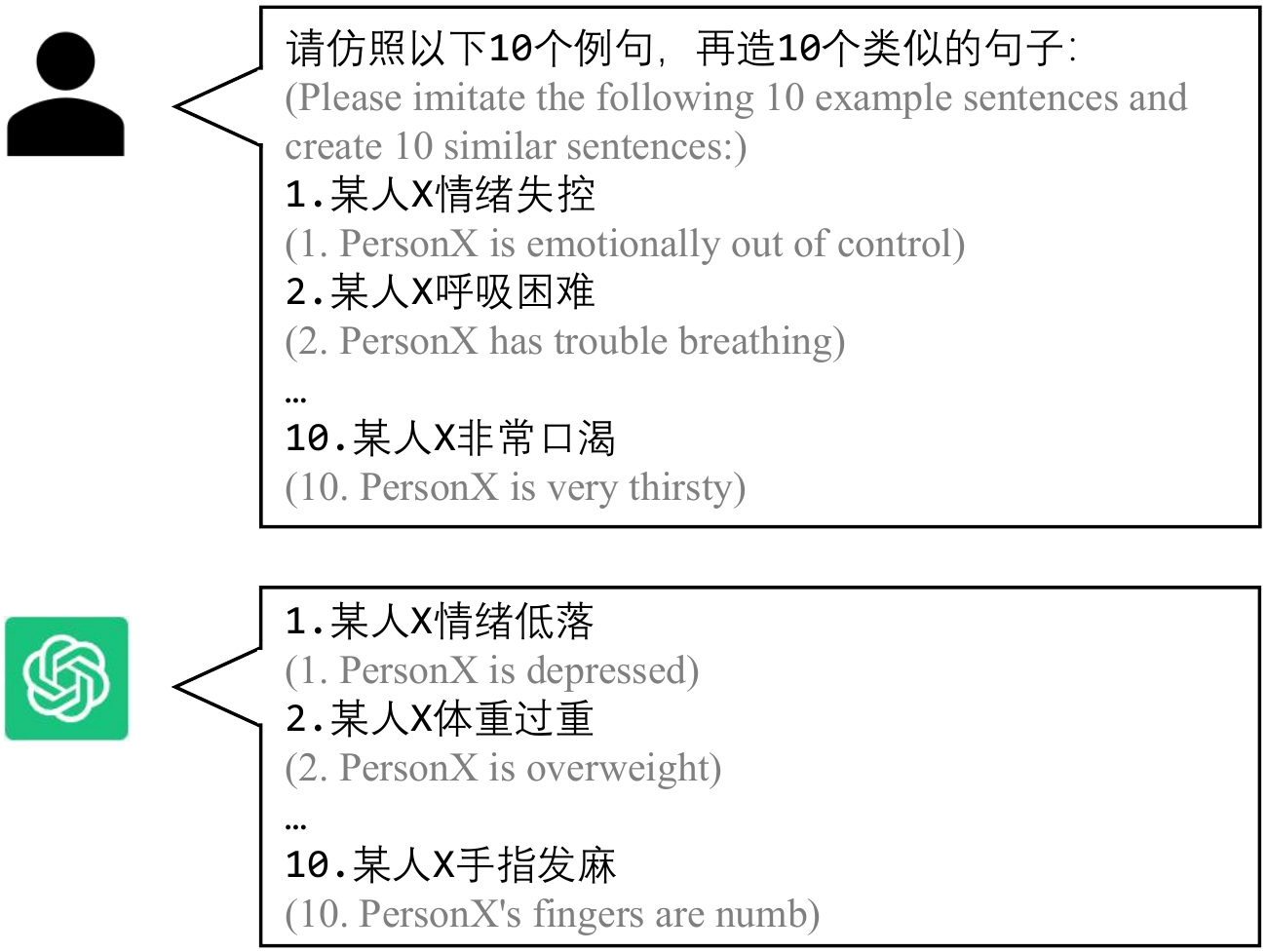}}
\caption{The prompt for distilling head items. \textcolor[RGB]{127,127,127}{Grey} indicates the translation which actually does not in the prompt.}
\label{fig:prompt_head_item}
\end{figure} 

\subsection{Collecting Head Items}
\label{subsec:2.1}

Following~\citet{wang-etal-2022-cn}, the head items of commonsense knowledge involve three knowledge types, \emph{i.e.}, voluntary items, involuntary items and state items.
Among them, voluntary items describe the events or activities that someone intentionally causes, such as ``PersonX learns to cook'', while involuntary items show the events or activities that someone is involuntarily involved in it, \emph{e.g.}, ``PersonX is attacked''. State items indicate the states that someone is in it for some time.

For each knowledge type, we manually create 200 head items by employing Chinese students, who majored in computer science, as volunteers.
The created head items are further checked by a data expert to ensure their quality.
Next, head items of each knowledge type serve as seeds to prompt ChatGPT to generate a large number of head items with the same type.
During the generation of ChatGPT, we utilize the official APIs provided by OpenAI.\footnote{\url{https://platform.openai.com/docs/guides/gpt/chat-completions-api}} To make a trade-off between the diversity and the generation quality, we set the temperature hyper-parameter to 0.7 based on the pilot experiments. The utilized prompt is shown in Figure~\ref{fig:prompt_head_item}, where a total of 10 example head items are randomly selected from the whole 200 seeds for each generation cycle.
Finally, we distill 61.7K unique head items from ChatGPT for each knowledge type, resulting in a total of 185.1K head items.

\begin{table*}[t]
\centering
\resizebox{0.95\textwidth}{!}
{
\begin{tabular}{lcccc}
\toprule[1pt]
\multicolumn{1}{c}{\multirow{2}{*}{Relation}} & \multirow{2}{*}{Explanation}                                             & \multicolumn{3}{c}{Valid Types of Head   Items} \\ 
\multicolumn{1}{c}{}                          &                                                                          & Voluntary       & Involuntary      & State      \\ \midrule[1pt]
xWant                                         & After the occurrence of \{Head\}, PersonX wants \{Tail\}                 & \usym{2714}               & \usym{2714}                & \usym{2714}          \\
xReact                                        & After the occurrence of \{Head\}, PersonX feels \{Tail\}                 & \usym{2714}               & \usym{2714}                & \usym{2717}          \\
xEffect                                       & After the occurrence of \{Head\}, PersonX does \{Tail\} as a   result    & \usym{2714}               & \usym{2714}               &  \usym{2714}         \\
xAttr                                         & After the occurrence of \{Head\}, we can know that PersonX is   \{Tail\} &  \usym{2714}              & \usym{2714}                & \usym{2714}         \\
xNeed                                         & Before the occurrence of \{Head\}, PersonX needs \{Tail\}                & \usym{2714}               & \usym{2714}                & \usym{2714}          \\
xIntent                                       & When doing \{Head\}, PersonX's intent is \{Tail\}                        & \usym{2714}               & \usym{2717}                &  \usym{2717}          \\
HinderedBy                                    & The occurrence of \{Head\} can be hindered by \{Tail\}                   & \usym{2714}               & \usym{2714}                & \usym{2714}         \\ \bottomrule[1pt]
\end{tabular}
}
\caption{The pre-defined relations during the construction of Snowman. ``\emph{\{Head\}}'' and ``\emph{\{Tail\}}'' denote the head and tail items, respectively. ``\emph{Valid Types of Head Items}'' indicates the relation is used to distill tail items for which types of head items.}
\label{table:predefined_relations}
\end{table*}

\begin{figure}[t]
\centerline{\includegraphics[width=0.45\textwidth]{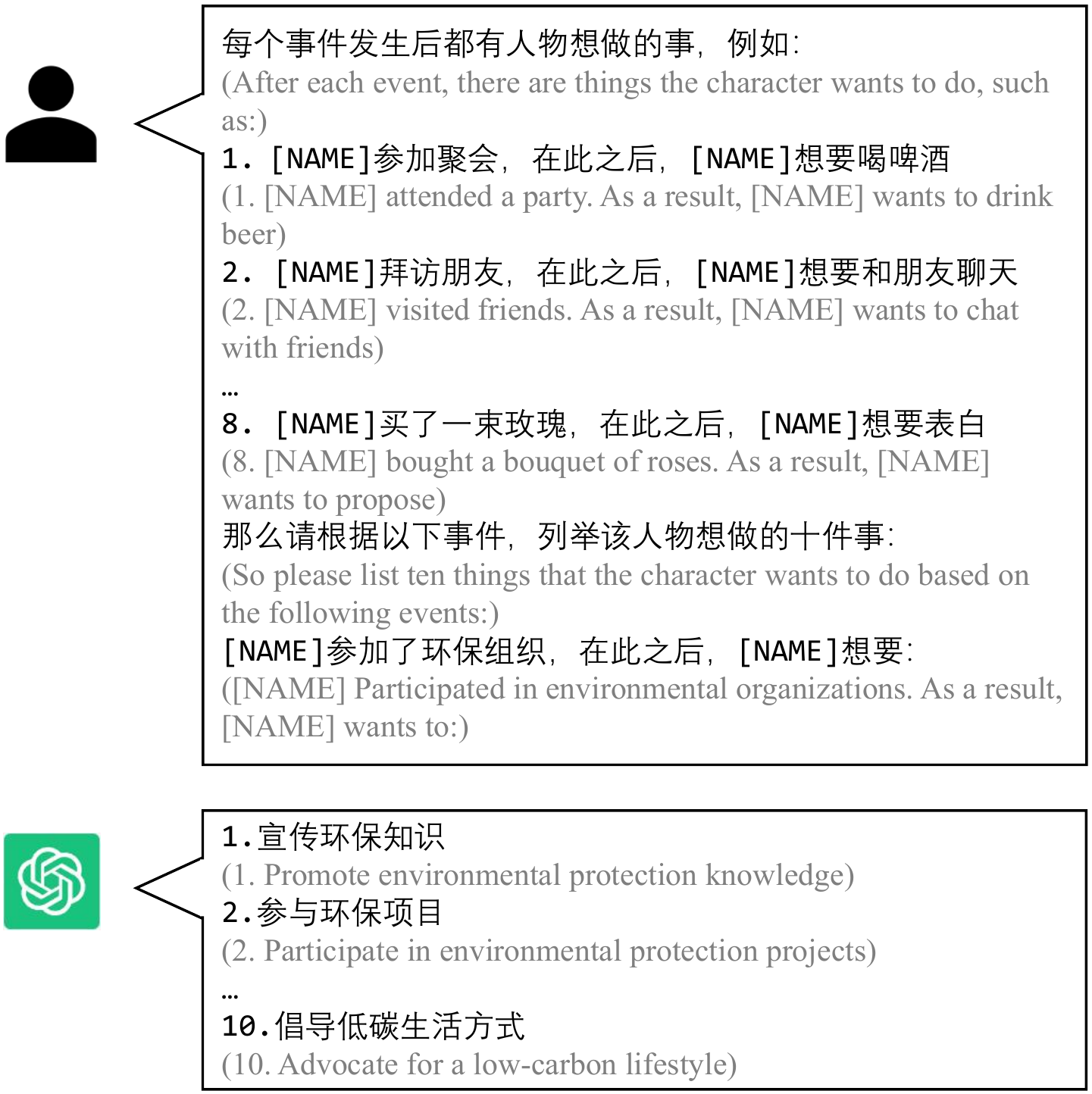}}
\caption{The prompt for distilling tail items. \textcolor[RGB]{127,127,127}{Grey} indicates the translation which actually does not in the prompt.}
\label{fig:prompt_tail_item}
\vspace{-0.5cm}
\end{figure}

\subsection{Collecting Tail Items}
\label{subsec:2.2}

After obtaining head items, we need to collect relations as well as tail items to get complete triples. Inspired by previous CKG construction work~\cite{ATOMIC,Hwang2020COMETATOMIC2O,wang-etal-2022-cn}, we choose seven mainstream relations, \emph{i.e.}, \emph{xWant}, \emph{xReact}, \emph{xEffect}, \emph{xAttr}, \emph{xNeed}, \emph{xIntent} and \emph{HinderedBy}. Table~\ref{table:predefined_relations} shows the explanation of each relation and the corresponding valid knowledge types of head items are used to distill tail items.
It is worth noting that each relation is only paired with the head items whose knowledge type is valid (c.f., ``\usym{2714}'' in Table~\ref{table:predefined_relations}) to avoid invalid distilling results. For example, the occurrence of an involuntary head item does not reflect any intent of someone due to his/her involuntary participation.

To distill ChatGPT to generate tail items, we also manually create 100 triple seeds with respect to each relation in the same way as Section~\ref{subsec:2.1}.
The prompt used to distill tail items is illustrated in Figure~\ref{fig:prompt_tail_item}, where the ``[NAME]'' placeholder is replaced with random Chinese common-used names. There are eight example triples in the prompt, and each of which is randomly selected from the seed triples. At the end of the prompt, we give a template sentence that describes the head item and the relation, and use it to distill ChatGPT to generate ten tail items.
In this manner, we finally obtain 11.1M unique Chinese commonsense triples, costing thousands of dollars to call official APIs.

\subsection{Self-Instruct Filtering}
\label{subsec:2.3}

To understand the quality of the distilled triples w.r.t each relation, we give a preliminary analysis (for more details, please refer to \S~\ref{sec:3}), and find that the human acceptance of triples with the \emph{HinderedBy} relation (65\%) is lower than that of others (typically more than 85\%). This is because the \emph{HinderedBy} triples describe the commonsense knowledge in negative statements, while most of the world knowledge exists in a positive form~\cite{molnar2000truthmakers,barker2012being}. Recent studies~\cite{Arnaout2023CanLL,Chen2023SayWY} also show that the quality of negative knowledge distilled from foundation models (even ChatGPT) still does not satisfy people.

To solve the above issue, we utilize a simple yet effective self-instruct filtering strategy. In detail, we sample 4K \emph{HinderedBy} triples and employ ChatGPT to judge whether the triple is valid.
Though ChatGPT struggles to generate \emph{HinderedBy} triples, it can well judge whether a \emph{HinderedBy} triple is valid. This is because the discriminative task is much easier than the generative counterpart.
Then, the boolean data is used to train a Chinese RoBERTa\footnote{\url{https://hf.co/hfl/chinese-roberta-wwm-ext-large}}~\cite{Liu2019RoBERTaAR} filter to filter low-quality raw \emph{HinderedBy} triples.
Note that this filtering method leverages the judgments of ChatGPT to filter out its low-quality generation, which does not need any human annotation and thus is called self-instruct filtering.

\section{Data Analysis}
\label{sec:3}

\begin{table*}[t]
\centering
\resizebox{0.95\textwidth}{!}
{
\begin{tabular}{lcccccc}
\toprule[1pt]
\multicolumn{1}{c}{\multirow{2}{*}{Commonsense Know Graph}} & \multirow{2}{*}{Language} & \multirow{2}{*}{Construction} & Unique     & Unique     & \multirow{2}{*}{Triples} & Human       \\
\multicolumn{1}{c}{}                                        &                           &                               & Head Items & Tail Items &                          & Acceptance  \\ \midrule[1pt]
ATOMIC 2020~\cite{Hwang2020COMETATOMIC2O}                                                & English                   & Crowdsourcing                         & 25,807     & 354,777    & 760,034                  & 86.8$^{*}$        \\
ATOMIC 10X$_{raw}$~\cite{west2021symbolic}                                     & English                   & Generation                    & 165,783    & 874,417    & 6,456,300                & 78.5$^{*}$        \\
ATOMIC 10X$_{high}$~\cite{west2021symbolic}                                   & English                   & Generation                    & 164,553    & 357,761    & 2,512,720                & 96.4$^{*}$        \\ \midrule[1pt]
ATOMIC-zh~\cite{li-etal-2022-c3kg}                                                   & Chinese                   & Translation                   & 20,949     & 276,446    & 712,970                  & 41.9 (38.7$^{\dagger}$) \\
CN-AutoMIC$_{raw}$~\cite{wang-etal-2022-cn}                                     & Chinese                   & Generation                    & 114,364    & 1,101,556  & 6,868,766                &  52.3 (47.6$^{\dagger}$) \\
CN-AutoMIC$_{high}$~\cite{wang-etal-2022-cn}                                           & Chinese                   & Generation                    & 89,738     & 182,893    & 1,140,840                &  89.5 (87.2$^{\dagger}$) \\
Snowman$_{raw}$ (Our)                                                     & Chinese                   & Generation                    & 185,075    & 5,783,395  & 11,087,873               & 86.8       \\
Snowman$_{high}$ (Our)                                                     & Chinese                   & Generation                    & 185,075    & 5,426,778   &  10,463,219    & 90.6      \\ \bottomrule[1pt]
\end{tabular}
}
\caption{The data statistics of Snowman and previous CKGs. $^{*}$ and $^{\dagger}$ denote the results are from \citet{west2021symbolic} and \citet{wang-etal-2022-cn}, respectively.}
\label{table:statistics}
\end{table*}

\noindent \textbf{Human Evaluation.}
We conduct human analysis on our Snowman and previous ATOMIC-zh~\cite{li-etal-2022-c3kg} as well as CN-AutoMIC~\cite{wang-etal-2022-cn}.
For Snowman, we randomly select 100 triples w.r.t each relation, resulting in 800 random triples (since the filtering strategy is applied in \emph{HinderedBy} relation, we randomly select 100 samples from unfiltered and filtered \emph{HinderedBy} triples, respectively).
For ATOMIC-zh and CN-AutoMIC, we randomly select 700 samples from each of them to conduct human analysis.
Following~\citet{wang-etal-2022-cn}, we invite three Chinese postgraduate students to assess whether each commonsense knowledge triple is reasonable, and compute the average proportion of reasonable triples for each CKG as its human acceptance.

\vspace{0.5ex}
\noindent \textbf{Overall Statistics.}
Table~\ref{table:statistics} compares the data statistics of our Snowman and previous CKGs. We find that the scale of Snowman is the largest among all CKGs, and only Snowman involves more than 10 million commonsense triples.
This finding indicates the superiority of ChatGPT serving as the backbone model to distill commonsense knowledge.
In addition, the human acceptance of raw Snowman reaches 86.8\%, which is higher than that of other unfiltered CKGs (78.5\% of ATOMIC 10X$_{raw}$ and 47.6\% of CN-AutoMIC$_{raw}$).
With the help of the self-instruct filtering strategy, the human acceptance of Snowman further reaches 90.6\%, the highest acceptance among all Chinese CKGs, indicating the effectiveness of the strategy.

\vspace{0.5ex}
\noindent \textbf{Relation-Level Statistics.} Table~\ref{table:relation_statistics} shows the relation-level data statistics of Snowman. We can find that the raw human acceptance of \emph{HinderedBy} relation only achieves 65.7\%, which is significantly lower than that of other relations (typically more than 85\%).
After filtering the low-quality \emph{HinderedBy} triples, the acceptance improves to 92.7\%, satisfying human evaluators.

\begin{table}[t]
\centering
\resizebox{0.45\textwidth}{!}
{
\begin{tabular}{lrrc}
\toprule[1pt]
\multicolumn{1}{c}{\multirow{2}{*}{Relation}} & \multicolumn{1}{c}{Unique}     & \multicolumn{1}{c}{\multirow{2}{*}{Triples}} & \multirow{2}{*}{Acceptance} \\
\multicolumn{1}{c}{}                          & \multicolumn{1}{c}{Tail Items} & \multicolumn{1}{c}{}                         &                             \\ \midrule[1pt]
xWant     &   1,280,417        &    1,850,336                                          & 89.3                          \\
xReact       &   206,947     &    1,227,908     & 93.0                          \\
xEffect           &      1,136,804          &      1,850,490     & 95.7                        \\
xAttr         &     389,639     &     1,846,849   & 87.3                        \\
xNeed       &    1,408,147    &  1,846,947      & 86.3                        \\
xIntent          &    317,298     &   616,821             & 90.0                          \\
HinderedBy (raw)   &   1,168,014     &   1,848,522             & 65.7                        \\
HinderedBy        &   810,754   &     1,223,868             & 92.7    \\ \bottomrule[1pt]                     
\end{tabular}
}
\caption{Relation-level data statistics of Snowman.}
\label{table:relation_statistics}
\end{table}

\section{Commonsense Generation}

\subsection{Experimental Setup}

To further demonstrate the usability and effectiveness of our Snowman, we conduct experiments on commonsense knowledge generation.
%
Specifically, following~\citet{wang-etal-2022-cn}, we train COMET model~\cite{Bosselut2019COMETCT} (a widely-used commonsense knowledge generation model) with the backbone of mT5-base (580M)\footnote{\url{https://hf.co/google/mt5-base}}~\cite{xue-etal-2021-mt5} on Snowman triples.
During training, we set 1e-4 learning rate, 128 batch size, and 2K warmup steps.
Next, we compare the trained model with other COMET models trained on ATOMIC-zh triples and CN-AutoMIC triples.
For each CKG, we only randomly select 500K triples to train the COMET model to avoid the impact of different data scales.
The test set is based on a held-out set (500 samples) of ATOMIC-zh for a fair comparison.
We manually evaluate the generation results in the same way as Section~\ref{sec:3} (\emph{i.e.}, human acceptance).

\begin{table}[t]
\centering
\resizebox{0.35\textwidth}{!}
{
\begin{tabular}{lc}
\toprule[1pt]
\multicolumn{1}{c}{Model} & Acceptance \\ \midrule[1pt]
COMET (ATOMIC-zh)         & 31.2\%    \\
COMET (CN-AutoMIC$_{raw}$)        & 47.8\%    \\
COMET (CN-AutoMIC$_{high}$)        & 61.3\%    \\
COMET (Snowman$_{low}$)           & 77.5\%    \\
COMET (Snowman$_{high}$)           & 81.8\%    \\ \bottomrule[1pt]
\end{tabular}
}
\caption{The performance of commonsense knowledge model trained on different CKGs.}
\label{table:comet_results}
\end{table}

\subsection{Results \& Discussion}
As shown in Table~\ref{table:comet_results}, the model trained on Snowman$_{high}$ achieves the best human acceptance, indicating the triples from our Snowman could help the generation model infer better commonsense knowledge.
This finding also demonstrates the high quality of our CKG and the superiority of the ChatGPT foundation model.
The instruction tuning and RLHF make the foundation models more intelligent, benefiting various downstream tasks and letting the construction of high-quality CKG become more convenient.

Compared with the previous CN-AutoMIC (that is distilled from mT5), we show that the cutting-edge foundation model (ChatGPT) can distill higher-quality Chinese CKG with only a few (hundred-level) seeds.
Besides, the construction protocol of Snowman is more concise than that of CN-AutoMIC (which includes cascaded filters and several rule-based denoising methods).
The construction protocol could also be trivially adapted to another language, and prompt the research on low-language CKGs as well as multi-lingual CKGs.

\section{Conclusion}

In this work, we transfer the success of foundation models to the construction of commonsense knowledge graphs (CKGs). In detail, we distill ChatGPT to generate Chinese commonsense triples with only a small number of seed items and triples. To deal with the negative knowledge issue, we design a simple yet effective self-instruct filtering strategy to filter out low-quality negative knowledge triples. Finally, our Snowman involves more than 10 million Chinese knowledge triples, making it the largest Chinese CKG. The human evaluation as well as commonsense generation experiments show the high quality of the constructed CKG.

\bibliography{anthology}

\begin{thebibliography}{46}
\expandafter\ifx\csname natexlab\endcsname\relax\def\natexlab#1{#1}\fi

\bibitem[{Ammanabrolu et~al.(2021)Ammanabrolu, Cheung, Broniec, and
  Riedl}]{Ammanabrolu_Cheung_Broniec_Riedl_2021}
Prithviraj Ammanabrolu, Wesley Cheung, William Broniec, and Mark~O. Riedl.
  2021.
\newblock \href {https://doi.org/10.1609/aaai.v35i7.16733} {Automated
  storytelling via causal, commonsense plot ordering}.
\newblock \emph{Proceedings of the AAAI Conference on Artificial Intelligence},
  35(7):5859--5867.

\bibitem[{Arnaout and Razniewski(2023)}]{Arnaout2023CanLL}
Hiba Arnaout and Simon Razniewski. 2023.
\newblock Can large language models generate salient negative statements?
\newblock \emph{ArXiv}, abs/2305.16755.

\bibitem[{Barker and Jago(2012)}]{barker2012being}
Stephen Barker and Mark Jago. 2012.
\newblock Being positive about negative facts.
\newblock \emph{Philosophy and Phenomenological research}, pages 117--138.

\bibitem[{Bosselut et~al.(2019)Bosselut, Rashkin, Sap, Malaviya, Celikyilmaz,
  and Choi}]{Bosselut2019COMETCT}
Antoine Bosselut, Hannah Rashkin, Maarten Sap, Chaitanya Malaviya, Asli
  Celikyilmaz, and Yejin Choi. 2019.
\newblock Comet: Commonsense transformers for automatic knowledge graph
  construction.
\newblock In \emph{Annual Meeting of the Association for Computational
  Linguistics}.

\bibitem[{Brown et~al.(2020)Brown, Mann, Ryder, Subbiah, Kaplan, Dhariwal,
  Neelakantan, Shyam, Sastry, Askell et~al.}]{brown2020language}
Tom Brown, Benjamin Mann, Nick Ryder, Melanie Subbiah, Jared~D Kaplan, Prafulla
  Dhariwal, Arvind Neelakantan, Pranav Shyam, Girish Sastry, Amanda Askell,
  et~al. 2020.
\newblock Language models are few-shot learners.
\newblock \emph{Advances in neural information processing systems},
  33:1877--1901.

\bibitem[{Bubeck et~al.(2023)Bubeck, Chandrasekaran, Eldan, Gehrke, Horvitz,
  Kamar, Lee, Lee, Li, Lundberg et~al.}]{bubeck2023sparks}
S{\'e}bastien Bubeck, Varun Chandrasekaran, Ronen Eldan, Johannes Gehrke, Eric
  Horvitz, Ece Kamar, Peter Lee, Yin~Tat Lee, Yuanzhi Li, Scott Lundberg,
  et~al. 2023.
\newblock Sparks of artificial general intelligence: Early experiments with
  gpt-4.
\newblock \emph{arXiv preprint arXiv:2303.12712}.

\bibitem[{Chen et~al.(2023)Chen, Shi, Fu, Cheng, Li, and Xiao}]{Chen2023SayWY}
Jiangjie Chen, Wei Shi, Ziquan Fu, Sijie Cheng, Lei Li, and Yanghua Xiao. 2023.
\newblock Say what you mean! large language models speak too positively about
  negative commonsense knowledge.
\newblock \emph{ArXiv}, abs/2305.05976.

\bibitem[{Feigenbaum(1984)}]{feigenbaum1984knowledge}
Edward~A Feigenbaum. 1984.
\newblock Knowledge engineering.
\newblock \emph{Annals of the New York Academy of Sciences}, 426(1):91--107.

\bibitem[{Hossain et~al.(2022)Hossain, Chinnappa, and
  Blanco}]{hossain-etal-2022-analysis}
Md~Mosharaf Hossain, Dhivya Chinnappa, and Eduardo Blanco. 2022.
\newblock \href {https://doi.org/10.18653/v1/2022.acl-short.81} {An analysis of
  negation in natural language understanding corpora}.
\newblock In \emph{Proceedings of the 60th Annual Meeting of the Association
  for Computational Linguistics (Volume 2: Short Papers)}, pages 716--723,
  Dublin, Ireland. Association for Computational Linguistics.

\bibitem[{Hwang et~al.(2020)Hwang, Bhagavatula, Bras, Da, Sakaguchi, Bosselut,
  and Choi}]{Hwang2020COMETATOMIC2O}
Jena~D. Hwang, Chandra Bhagavatula, Ronan~Le Bras, Jeff Da, Keisuke Sakaguchi,
  Antoine Bosselut, and Yejin Choi. 2020.
\newblock Comet-atomic 2020: On symbolic and neural commonsense knowledge
  graphs.
\newblock In \emph{AAAI Conference on Artificial Intelligence}.

\bibitem[{Lenat(1995)}]{lenat1995cyc}
Douglas~B Lenat. 1995.
\newblock Cyc: A large-scale investment in knowledge infrastructure.
\newblock \emph{Communications of the ACM}, 38(11):33--38.

\bibitem[{Li et~al.(2022)Li, Li, Zhang, Li, Wei, Cui, and
  Wang}]{li-etal-2022-c3kg}
Dawei Li, Yanran Li, Jiayi Zhang, Ke~Li, Chen Wei, Jianwei Cui, and Bin Wang.
  2022.
\newblock \href {https://doi.org/10.18653/v1/2022.findings-acl.107}
  {{C}$^3${KG}: A {C}hinese commonsense conversation knowledge graph}.
\newblock In \emph{Findings of the Association for Computational Linguistics:
  ACL 2022}, pages 1369--1383, Dublin, Ireland. Association for Computational
  Linguistics.

\bibitem[{Liang et~al.(2023)Liang, Meng, Wang, Xu, Chen, and Zhou}]{liang2023d}
Yunlong Liang, Fandong Meng, Jiaan Wang, Jinan Xu, Yufeng Chen, and Jie Zhou.
  2023.
\newblock D2tv: Dual knowledge distillation and target-oriented vision modeling
  for many-to-many multimodal summarization.
\newblock \emph{arXiv preprint arXiv:2305.12767}.

\bibitem[{Liang et~al.(2022{\natexlab{a}})Liang, Meng, Xu, Wang, Chen, and
  Zhou}]{liang2022summary}
Yunlong Liang, Fandong Meng, Jinan Xu, Jiaan Wang, Yufeng Chen, and Jie Zhou.
  2022{\natexlab{a}}.
\newblock Summary-oriented vision modeling for multimodal abstractive
  summarization.
\newblock \emph{arXiv preprint arXiv:2212.07672}.

\bibitem[{Liang et~al.(2021)Liang, Meng, Zhang, Chen, Xu, and
  Zhou}]{liang2020infusing}
Yunlong Liang, Fandong Meng, Ying Zhang, Yufeng Chen, Jinan Xu, and Jie Zhou.
  2021.
\newblock \href {https://ojs.aaai.org/index.php/AAAI/article/view/17575}
  {Infusing multi-source knowledge with heterogeneous graph neural network for
  emotional conversation generation}.
\newblock \emph{Proceedings of AAAI}, pages 13343--13352.

\bibitem[{Liang et~al.(2022{\natexlab{b}})Liang, Meng, Zhang, Chen, Xu, and
  Zhou}]{liang2022emotional}
Yunlong Liang, Fandong Meng, Ying Zhang, Yufeng Chen, Jinan Xu, and Jie Zhou.
  2022{\natexlab{b}}.
\newblock Emotional conversation generation with heterogeneous graph neural
  network.
\newblock \emph{Artificial Intelligence}, 308:103714.

\bibitem[{Liang et~al.(2022{\natexlab{c}})Liang, Meng, Zhou, Xu, Chen, Su, and
  Zhou}]{liang-etal-2022-variational}
Yunlong Liang, Fandong Meng, Chulun Zhou, Jinan Xu, Yufeng Chen, Jinsong Su,
  and Jie Zhou. 2022{\natexlab{c}}.
\newblock \href {https://doi.org/10.18653/v1/2022.acl-long.148} {A variational
  hierarchical model for neural cross-lingual summarization}.
\newblock In \emph{ACL}, pages 2088--2099.

\bibitem[{Liu et~al.(2019)Liu, Ott, Goyal, Du, Joshi, Chen, Levy, Lewis,
  Zettlemoyer, and Stoyanov}]{Liu2019RoBERTaAR}
Yinhan Liu, Myle Ott, Naman Goyal, Jingfei Du, Mandar Joshi, Danqi Chen, Omer
  Levy, Mike Lewis, Luke Zettlemoyer, and Veselin Stoyanov. 2019.
\newblock Roberta: A robustly optimized bert pretraining approach.
\newblock \emph{ArXiv}, abs/1907.11692.

\bibitem[{Molnar(2000)}]{molnar2000truthmakers}
George Molnar. 2000.
\newblock Truthmakers for negative truths.
\newblock \emph{Australasian Journal of philosophy}, 78(1):72--86.

\bibitem[{Mostafazadeh et~al.(2020)Mostafazadeh, Kalyanpur, Moon, Buchanan,
  Berkowitz, Biran, and Chu-Carroll}]{mostafazadeh-etal-2020-glucose}
Nasrin Mostafazadeh, Aditya Kalyanpur, Lori Moon, David Buchanan, Lauren
  Berkowitz, Or~Biran, and Jennifer Chu-Carroll. 2020.
\newblock \href {https://doi.org/10.18653/v1/2020.emnlp-main.370} {{GLUCOSE}:
  {G}enera{L}ized and {CO}ntextualized story explanations}.
\newblock In \emph{Proceedings of the 2020 Conference on Empirical Methods in
  Natural Language Processing (EMNLP)}, pages 4569--4586, Online. Association
  for Computational Linguistics.

\bibitem[{Muennighoff et~al.(2022)Muennighoff, Wang, Sutawika, Roberts,
  Biderman, Scao, Bari, Shen, Yong, Schoelkopf, Tang, Radev, Aji, Almubarak,
  Albanie, Alyafeai, Webson, Raff, and Raffel}]{Muennighoff2022CrosslingualGT}
Niklas Muennighoff, Thomas Wang, Lintang Sutawika, Adam Roberts, Stella~Rose
  Biderman, Teven~Le Scao, M~Saiful Bari, Sheng Shen, Zheng~Xin Yong, Hailey
  Schoelkopf, Xiangru Tang, Dragomir~R. Radev, Alham~Fikri Aji, Khalid
  Almubarak, Samuel Albanie, Zaid Alyafeai, Albert Webson, Edward Raff, and
  Colin Raffel. 2022.
\newblock Crosslingual generalization through multitask finetuning.
\newblock \emph{ArXiv}, abs/2211.01786.

\bibitem[{Nguyen et~al.(2021)Nguyen, Razniewski, and Weikum}]{advanced}
Tuan-Phong Nguyen, Simon Razniewski, and Gerhard Weikum. 2021.
\newblock \href {https://doi.org/10.1145/3442381.3449827} {Advanced semantics
  for commonsense knowledge extraction}.
\newblock In \emph{Proceedings of the Web Conference 2021}, WWW '21, page
  2636–2647, New York, NY, USA. Association for Computing Machinery.

\bibitem[{OpenAI(2022)}]{ChatGPT}
OpenAI. 2022.
\newblock Introducing chatgpt.
\newblock \url{https://openai.com/blog/chatgpt}.

\bibitem[{OpenAI(2023)}]{OpenAI2023GPT4TR}
OpenAI. 2023.
\newblock Gpt-4 technical report.
\newblock \emph{ArXiv}, abs/2303.08774.

\bibitem[{Raffel et~al.(2020)Raffel, Shazeer, Roberts, Lee, Narang, Matena,
  Zhou, Li, and Liu}]{raffel2020exploring}
Colin Raffel, Noam Shazeer, Adam Roberts, Katherine Lee, Sharan Narang, Michael
  Matena, Yanqi Zhou, Wei Li, and Peter~J Liu. 2020.
\newblock Exploring the limits of transfer learning with a unified text-to-text
  transformer.
\newblock \emph{The Journal of Machine Learning Research}, 21(1):5485--5551.

\bibitem[{Romero et~al.(2019)Romero, Razniewski, Pal, Z.~Pan, Sakhadeo, and
  Weikum}]{10.1145/3357384.3357955}
Julien Romero, Simon Razniewski, Koninika Pal, Jeff Z.~Pan, Archit Sakhadeo,
  and Gerhard Weikum. 2019.
\newblock \href {https://doi.org/10.1145/3357384.3357955} {Commonsense
  properties from query logs and question answering forums}.
\newblock In \emph{Proceedings of the 28th ACM International Conference on
  Information and Knowledge Management}, CIKM '19, page 1411–1420, New York,
  NY, USA. Association for Computing Machinery.

\bibitem[{Sap et~al.(2019)Sap, Le~Bras, Allaway, Bhagavatula, Lourie, Rashkin,
  Roof, Smith, and Choi}]{ATOMIC}
Maarten Sap, Ronan Le~Bras, Emily Allaway, Chandra Bhagavatula, Nicholas
  Lourie, Hannah Rashkin, Brendan Roof, Noah~A. Smith, and Yejin Choi. 2019.
\newblock \href {https://doi.org/10.1609/aaai.v33i01.33013027} {Atomic: An
  atlas of machine commonsense for if-then reasoning}.
\newblock \emph{Proceedings of the AAAI Conference on Artificial Intelligence},
  33(01):3027--3035.

\bibitem[{Schulman et~al.(2017)Schulman, Wolski, Dhariwal, Radford, and
  Klimov}]{Schulman2017ProximalPO}
John Schulman, Filip Wolski, Prafulla Dhariwal, Alec Radford, and Oleg Klimov.
  2017.
\newblock Proximal policy optimization algorithms.
\newblock \emph{ArXiv}, abs/1707.06347.

\bibitem[{Speer et~al.(2017)Speer, Chin, and Havasi}]{speer2017conceptnet}
Robyn Speer, Joshua Chin, and Catherine Havasi. 2017.
\newblock Conceptnet 5.5: An open multilingual graph of general knowledge.
\newblock In \emph{Proceedings of the AAAI conference on artificial
  intelligence}, volume~31.

\bibitem[{Stiennon et~al.(2020)Stiennon, Ouyang, Wu, Ziegler, Lowe, Voss,
  Radford, Amodei, and Christiano}]{stiennon2020learning}
Nisan Stiennon, Long Ouyang, Jeffrey Wu, Daniel Ziegler, Ryan Lowe, Chelsea
  Voss, Alec Radford, Dario Amodei, and Paul~F Christiano. 2020.
\newblock Learning to summarize with human feedback.
\newblock \emph{Advances in Neural Information Processing Systems},
  33:3008--3021.

\bibitem[{Tandon et~al.(2014)Tandon, de~Melo, Suchanek, and
  Weikum}]{10.1145/2556195.2556245}
Niket Tandon, Gerard de~Melo, Fabian Suchanek, and Gerhard Weikum. 2014.
\newblock \href {https://doi.org/10.1145/2556195.2556245} {Webchild: Harvesting
  and organizing commonsense knowledge from the web}.
\newblock In \emph{Proceedings of the 7th ACM International Conference on Web
  Search and Data Mining}, WSDM '14, page 523–532, New York, NY, USA.
  Association for Computing Machinery.

\bibitem[{Tian et~al.(2020)Tian, Zhang, Liu, Zhao, Jia, and
  Sheng}]{tian-etal-2020-scene}
Zhixing Tian, Yuanzhe Zhang, Kang Liu, Jun Zhao, Yantao Jia, and Zhicheng
  Sheng. 2020.
\newblock \href {https://doi.org/10.18653/v1/2020.emnlp-main.247} {Scene
  restoring for narrative machine reading comprehension}.
\newblock In \emph{Proceedings of the 2020 Conference on Empirical Methods in
  Natural Language Processing (EMNLP)}, pages 3063--3073, Online. Association
  for Computational Linguistics.

\bibitem[{Wang et~al.(2022{\natexlab{a}})Wang, Li, Chen, Liu, and
  Zhao}]{wang-etal-2022-cn}
Chenhao Wang, Jiachun Li, Yubo Chen, Kang Liu, and Jun Zhao.
  2022{\natexlab{a}}.
\newblock \href {https://aclanthology.org/2022.emnlp-main.628}
  {{CN}-{A}uto{MIC}: Distilling {C}hinese commonsense knowledge from pretrained
  language models}.
\newblock In \emph{Proceedings of the 2022 Conference on Empirical Methods in
  Natural Language Processing}, pages 9253--9265, Abu Dhabi, United Arab
  Emirates. Association for Computational Linguistics.

\bibitem[{Wang et~al.(2022{\natexlab{b}})Wang, Li, Zhang, Zheng, Qu, Liu, Zhao,
  and Chen}]{10.1145/3488560.3498405}
Jiaan Wang, Zhixu Li, Tingyi Zhang, Duo Zheng, Jianfeng Qu, An~Liu, Lei Zhao,
  and Zhigang Chen. 2022{\natexlab{b}}.
\newblock \href {https://doi.org/10.1145/3488560.3498405} {Knowledge enhanced
  sports game summarization}.
\newblock In \emph{Proceedings of the Fifteenth ACM International Conference on
  Web Search and Data Mining}, WSDM '22, page 1045–1053, New York, NY, USA.
  Association for Computing Machinery.

\bibitem[{Wang et~al.(2023{\natexlab{a}})Wang, Liang, Meng, Li, Qu, and
  Zhou}]{Wang2023CrossLingualSV}
Jiaan Wang, Yunlong Liang, Fandong Meng, Zhixu Li, Jianfeng Qu, and Jie Zhou.
  2023{\natexlab{a}}.
\newblock Cross-lingual summarization via chatgpt.
\newblock \emph{ArXiv}, abs/2302.14229.

\bibitem[{Wang et~al.(2023{\natexlab{b}})Wang, Liang, Meng, Shi, Li, Xu, Qu,
  and Zhou}]{Wang2023IsCA}
Jiaan Wang, Yunlong Liang, Fandong Meng, Haoxiang Shi, Zhixu Li, Jinan Xu,
  Jianfeng Qu, and Jie Zhou. 2023{\natexlab{b}}.
\newblock Is chatgpt a good nlg evaluator? a preliminary study.
\newblock \emph{ArXiv}, abs/2303.04048.

\bibitem[{Wang et~al.(2022{\natexlab{c}})Wang, Meng, Lu, Zheng, Li, Qu, and
  Zhou}]{Wang2022ClidSumAB}
Jiaan Wang, Fandong Meng, Ziyao Lu, Duo Zheng, Zhixu Li, Jianfeng Qu, and Jie
  Zhou. 2022{\natexlab{c}}.
\newblock \href {https://aclanthology.org/2022.emnlp-main.526} {{C}lid{S}um: A
  benchmark dataset for cross-lingual dialogue summarization}.
\newblock In \emph{Proceedings of the 2022 Conference on Empirical Methods in
  Natural Language Processing}, pages 7716--7729, Abu Dhabi, United Arab
  Emirates. Association for Computational Linguistics.

\bibitem[{Wang et~al.(2022{\natexlab{d}})Wang, Meng, Zhang, Liang, Xu, Li, and
  Zhou}]{Wang2022UnderstandingTI}
Jiaan Wang, Fandong Meng, Tingyi Zhang, Yunlong Liang, Jiarong Xu, Zhixu Li,
  and Jie Zhou. 2022{\natexlab{d}}.
\newblock Understanding translationese in cross-lingual summarization.
\newblock \emph{ArXiv}, abs/2212.07220.

\bibitem[{Wang et~al.(2022{\natexlab{e}})Wang, Meng, Zheng, Liang, Li, Qu, and
  Zhou}]{Wang2022ASO}
Jiaan Wang, Fandong Meng, Duo Zheng, Yunlong Liang, Zhixu Li, Jianfeng Qu, and
  Jie Zhou. 2022{\natexlab{e}}.
\newblock \href {https://doi.org/10.1162/tacl_a_00520} {{A Survey on
  Cross-Lingual Summarization}}.
\newblock \emph{Transactions of the Association for Computational Linguistics},
  10:1304--1323.

\bibitem[{Wang et~al.(2023{\natexlab{c}})Wang, Meng, Zheng, Liang, Li, Qu, and
  Zhou}]{Wang2023TowardsUM}
Jiaan Wang, Fandong Meng, Duo Zheng, Yunlong Liang, Zhixu Li, Jianfeng Qu, and
  Jie Zhou. 2023{\natexlab{c}}.
\newblock Towards unifying multi-lingual and cross-lingual summarization.
\newblock \emph{ArXiv}, abs/2305.09220.

\bibitem[{Wang et~al.(2022{\natexlab{f}})Wang, Zou, Li, Qu, Zhao, Liu, and
  Zhao}]{Wang2022IncorporatingCK}
Jiaan Wang, Beiqi Zou, Zhixu Li, Jianfeng Qu, Pengpeng Zhao, An~Liu, and Lei
  Zhao. 2022{\natexlab{f}}.
\newblock Incorporating commonsense knowledge into story ending generation via
  heterogeneous graph networks.
\newblock In \emph{International Conference on Database Systems for Advanced
  Applications}.

\bibitem[{Wei et~al.(2021)Wei, Bosma, Zhao, Guu, Yu, Lester, Du, Dai, and
  Le}]{wei2021finetuned}
Jason Wei, Maarten Bosma, Vincent~Y Zhao, Kelvin Guu, Adams~Wei Yu, Brian
  Lester, Nan Du, Andrew~M Dai, and Quoc~V Le. 2021.
\newblock Finetuned language models are zero-shot learners.
\newblock \emph{arXiv preprint arXiv:2109.01652}.

\bibitem[{West et~al.(2021)West, Bhagavatula, Hessel, Hwang, Jiang, Bras, Lu,
  Welleck, and Choi}]{west2021symbolic}
Peter West, Chandra Bhagavatula, Jack Hessel, Jena~D Hwang, Liwei Jiang,
  Ronan~Le Bras, Ximing Lu, Sean Welleck, and Yejin Choi. 2021.
\newblock Symbolic knowledge distillation: from general language models to
  commonsense models.
\newblock \emph{arXiv preprint arXiv:2110.07178}.

\bibitem[{Xue et~al.(2021)Xue, Constant, Roberts, Kale, Al-Rfou, Siddhant,
  Barua, and Raffel}]{xue-etal-2021-mt5}
Linting Xue, Noah Constant, Adam Roberts, Mihir Kale, Rami Al-Rfou, Aditya
  Siddhant, Aditya Barua, and Colin Raffel. 2021.
\newblock \href {https://doi.org/10.18653/v1/2021.naacl-main.41} {m{T}5: A
  massively multilingual pre-trained text-to-text transformer}.
\newblock In \emph{Proceedings of the 2021 Conference of the North American
  Chapter of the Association for Computational Linguistics: Human Language
  Technologies}, pages 483--498, Online. Association for Computational
  Linguistics.

\bibitem[{Yu et~al.(2022)Yu, Chatterjee, Asai, Hu, and
  Choi}]{yu-etal-2022-beyond}
Xinyan Yu, Trina Chatterjee, Akari Asai, Junjie Hu, and Eunsol Choi. 2022.
\newblock \href {https://aclanthology.org/2022.findings-emnlp.273} {Beyond
  counting datasets: A survey of multilingual dataset construction and
  necessary resources}.
\newblock In \emph{Findings of the Association for Computational Linguistics:
  EMNLP 2022}, pages 3725--3743, Abu Dhabi, United Arab Emirates. Association
  for Computational Linguistics.

\bibitem[{Zhang et~al.(2020)Zhang, Khashabi, Song, and Roth}]{ijcai2020p554}
Hongming Zhang, Daniel Khashabi, Yangqiu Song, and Dan Roth. 2020.
\newblock \href {https://doi.org/10.24963/ijcai.2020/554} {Transomcs: From
  linguistic graphs to commonsense knowledge}.
\newblock In \emph{Proceedings of the Twenty-Ninth International Joint
  Conference on Artificial Intelligence, {IJCAI-20}}, pages 4004--4010.
  International Joint Conferences on Artificial Intelligence Organization.
\newblock Main track.

\end{thebibliography}
\bibliographystyle{acl_natbib}

\appendix



\end{document}